\documentclass[10pt,twocolumn,letterpaper]{article}

\usepackage{cvpr}
\usepackage{cvpr}
\usepackage{times}
\usepackage{epsfig}
\usepackage{graphicx}
\usepackage{amsmath}
\usepackage{amssymb}
\usepackage{comment}
\usepackage{textcomp, gensymb}
\usepackage{array}
\usepackage{tabularx}
\usepackage{enumitem}

\usepackage[table]{xcolor}

\begin{document}

\title{Recognizing People by Body Shape Using Deep Networks of Images and Words}


\author{Blake A. Myers\\
The University of Texas at Dallas\and   Lucas Jaggernauth\\
The University of Texas at Dallas\and  Thomas M. Metz\\
The University of Texas at Dallas\and  Matthew Q. Hill\\
The University of Texas at Dallas \and Veda Nandan Gandi\\
 The University of Texas at Dallas \and Carlos D. Castillo\\
Johns Hopkins University \and Alice J. O’Toole\\
The University of Texas at Dallas}

\maketitle
\thispagestyle{empty}

\begin{abstract}
   Common and important applications of person identification occur at distances and viewpoints in which the face is not visible or is not sufficiently resolved to be useful. We examine body shape as a biometric across distance and viewpoint variation. We propose an approach that combines standard object classification networks with representations based on linguistic (word-based) descriptions of bodies. Algorithms with and without linguistic training were compared on their ability to identify people from  body shape in images captured across a large range of distances/views (close-range, 100m, 200m, 270m, 300m, 370m, 400m, 490m, 500m, 600m, and at elevated pitch in images taken by an unmanned aerial vehicle [UAV]). Accuracy, as measured by identity-match ranking and false accept errors in an open-set test, was surprisingly good. For identity-ranking, linguistic models were more accurate for close-range images, whereas non-linguistic models fared better at intermediary distances. Fusion of the linguistic and non-linguistic embeddings improved performance at all, but the farthest distance. Although the non-linguistic model yielded fewer false accepts at all distances, fusion of the linguistic and non-linguistic models decreased false accepts for all, but the UAV images. We conclude that linguistic and non-linguistic representations of body shape can offer complementary identity information for bodies that can improve identification in applications of interest.     
   
\end{abstract}

\section{Introduction}

Studies of visually based person recognition have concentrated primarily on identity cues in the face because faces provide nearly unique information about identity. Consistent with this  axiom, behavioral studies indicate that when the entire person is visible, people rely strongly on the face \cite{robbins2012effects}, even when the body can be useful \cite{rice2013role}. However, it is important in some cases, to identify people from distances and viewpoints that limit the resolution or visibility of the face. These types of vantage points are common in natural viewing environments and are typical for surveillance cameras that monitor large areas of space. 

The use of bodies for identification in natural viewing conditions
is possible because body identity cues (e.g., shape, structure of body) remain visible and easy to resolve across a large range of distances. Concomitantly, behavioral studies indicate that people rely
on bodies when the face is not easy to see \cite{hahn2016dissecting} or provides poor or misleading information about identity \cite{rice2013role,rice2013unaware}. 
Although the identity cues from the body are inherently less diagnostic than cues from the face, they can nonetheless support surprisingly accurate identification for human perceivers, both in isolation \cite{hahn2016dissecting, rice2013role, rice2013unaware, robbins2012effects} and in combination with other visually robust biometrics
such as gait \cite{kale2004identification, yovel2016recognizing}.  Thus, the use of body identity information as a component of biometric fusion (e.g., face, body, and gait) offers the possibility of improving person recognition in challenging viewing conditions.

In this study, we  examined the extent to which  body shape can support accurate identification. We distinguish this problem from the more commonly studied {\it person re-identification} problem, which ``aims to retrieve a person of interest across multiple non-overlapping cameras'' (see \cite{ye2021deep} for a review). Although the body is used extensively in re-identification, many ``identity'' cues that operate at this short time scale (e.g., clothing, footwear, hairstyle, accessories) cannot be relied upon over a longer time period. In eliminating these short-term cues, the shape and structure of the body remain the primary cues to identity.

Person identification using body shape is challenging for multiple reasons. First, the human body  is a three-dimensional object capable of both rigid and non-rigid deformation. Bodies can appear from multiple views, with limbs in any number of configurations or poses (e.g., arms up, leg lifted).  
Second, changes in clothing can alter the color and texture of the body in unpredictable ways. This is a challenging problem in light of multiple findings indicating that the most commonly employed machine learning algorithms (i.e., deep convolutional neural networks [DCNNs]) show a strong texture bias in classification, with less sensitivity to the global shape of an object (cf. \cite{geirhos2018imagenet}). 

Third, high-quality datasets available for training body recognition algorithms across diverse viewpoints, distances, and appearances (e.g., clothing) are  quite limited. This makes it difficult to use training diversity 
to overcome the texture bias problem and other types of over-fitting.

Our approach combines elements of a standard
image-based deep network with a body representation trained to produce a description of the body based on a small number ($n=30$) of linguistic attributes (cf. \cite{hill2016creating, streuber2016body} and Table~\ref{descriptors}). 
Human-annotated linguistic descriptors (e.g., words such as curvy, pear-shaped, muscular, and broad shoulders) offer advantages for body shape 
representations, because they are accessible and constant over a wide range of viewing distances, angles, clothing, and footwear. 
 A ``curvy woman'' or
``muscular, broad-shouldered'' man can be perceived and  
annotated easily over photometric change. Moreover,
each word can entail multiple global and local anthropometric features. For example, the word  ``pear-shaped'' indicates large hips relative to bust size and suggests a bottom-heavy body shape.
Combinations of these words are even more powerful in conjuring up the gross shape of a body. We can imagine a tall, muscular, athletic, broad-shouldered man quite easily, and distinguish this body shape from other types of athletic, muscular bodies (e.g., cyclist, rock climber) \cite{hill2016creating, streuber2016body}.

\begin{table}
\begin{center}
\caption{Linguistic descriptors used to annotate frames.}\label{descriptors}
\vskip .25cm
\begin{tabular}{l  l  l }
proportioned & rectangular & stocky \\
short legs & muscular & average \\
tall & sturdy & big \\
long legs & lean & short torso \\
pear-shaped & petite & broad shoulders \\
heavy set & long & long torso \\
round (apple) & built & fit \\
skinny & masculine & small \\
pear-shaped & petite & broad shoulders \\
short & feminine & curvy\\
\end{tabular}
\vskip -0.9cm
\end{center}
\end{table}

The contributions of the paper are:
\begin{itemize}[nosep]
    \item Proposal of a body identification algorithm that uses a combination of object recognition strategies and linguistic body descriptions. 
    \item Direct comparisons between models pre-trained with and without a linguistic core.
    \item Evaluation of the algorithms across a broad range of distances (close-up, 100m, 200m, 270m, 300m, 370m, 400m, 490m, 500m, 600m), viewpoints, and at elevated pitches from a UAV.
    \item Demonstration that the performance of linguistic and non-linguistic models differs at different distances, and with UAV.    
    \item Demonstration that fusing linguistic and non-linguistic models can improve body identification over either linguistic or non-linguistic core models operating alone.     
    \item Novel body shape algorithms that operate in challenging testing conditions (mismatched clothing, long-distances, difficult viewing angles). 

\end{itemize}

\begin{figure}[t]
\begin{center}
   \includegraphics[width=\linewidth]{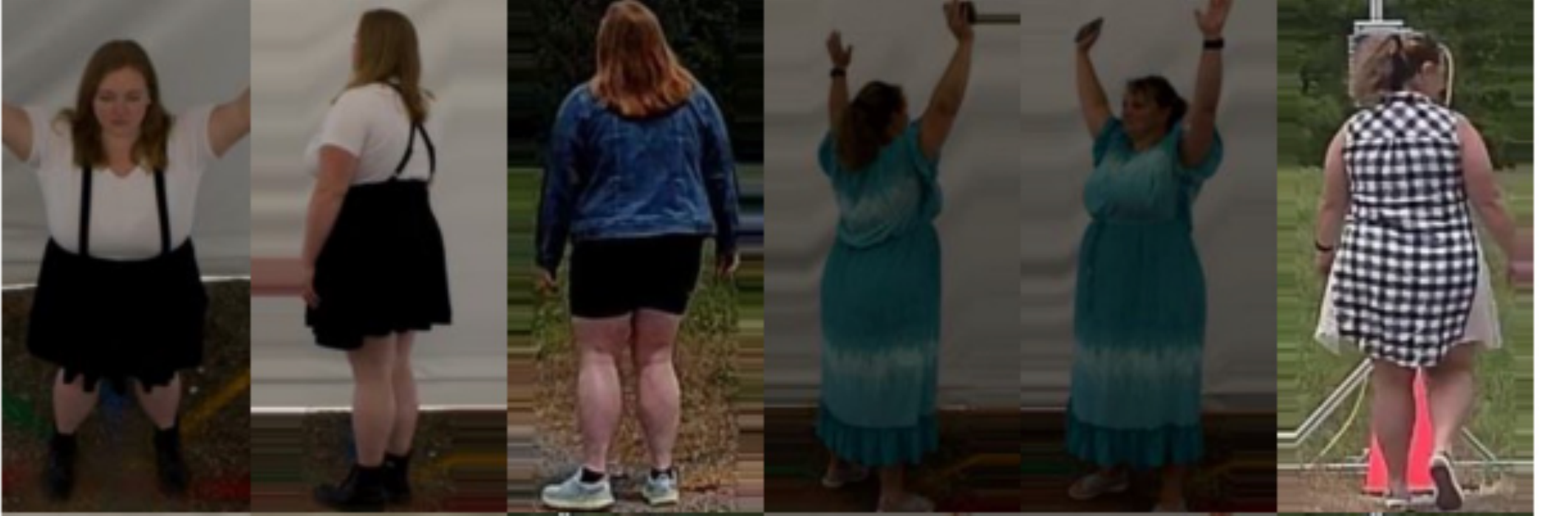}\\
   \includegraphics[width=\linewidth]{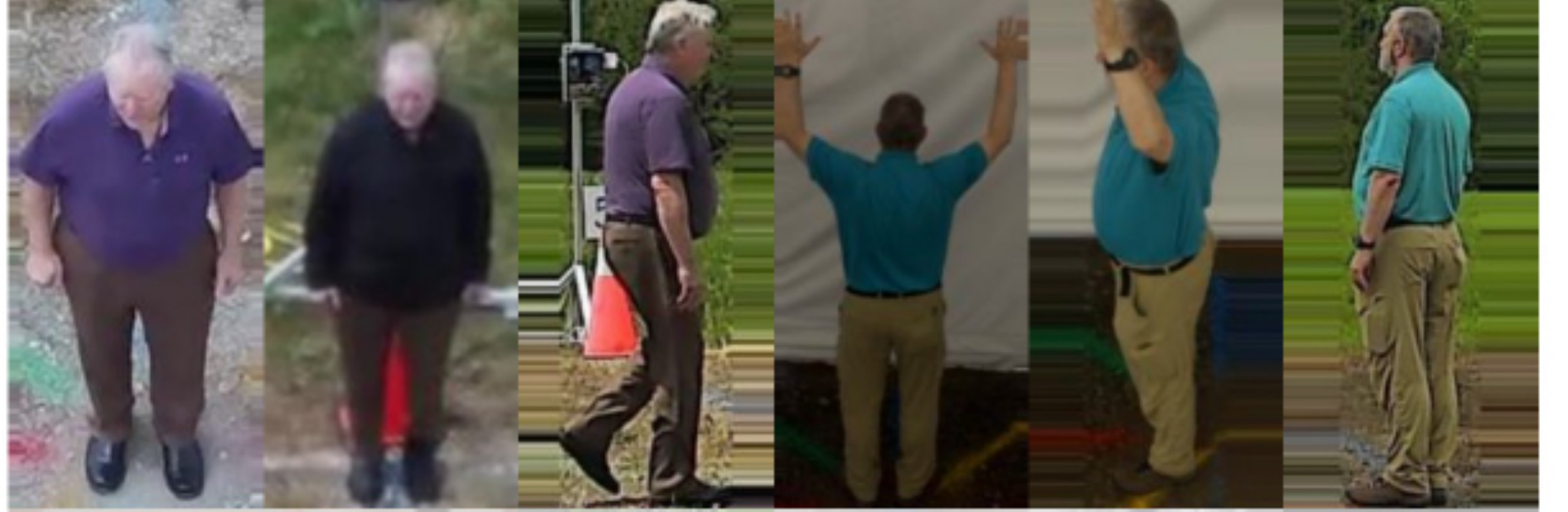}
\end{center}
\caption{Example image frames from the BRS dataset that show two identities from multiple viewpoints. }
\label{fig:BRS_examples}
\end{figure}

\subsection{Related Work}
\textbf{Body Shape Identification.}
Work using body shape as a biometric is surprisingly limited. In one early paper
\cite{godil2003human}, 73 human-placed anthropometric landmarks from bodies in the Civilian American and European Surface Anthropometry Resource [CAESAR] dataset
were used as a body shape representation. CAESAR is a database with 5000+ 3D laser scans of American and European adults. The utility of these landmarks for identity verification was tested for a gallery set of original anthropometric measures and a probe set consisting of noise-perturbed versions of the gallery. Although the model performed well,  the landmark measures are useful only when they are available with 3D body models.

In metric terms, body measures including height and weight have been used as body biometrics in forensic scenarios. 
In using a 3D model to estimate these
quantities, height and weight were estimated accurately for a wide range of body poses and camera angles when a 3D model  can be reliably determined \cite{thakkar2021feasibility}. However,  scale disambiguation was not feasible in real-world scenarios with coherent point drift (CPD) \cite{myronenko2010point}. Thus, the authors concluded that ``3D pose estimation'' is necessary, but not sufficient, to achieve accurate estimates of height and weight. 

In a subsequent study \cite{Thakkar_2022_CVPR},
 height and weight were estimated using pose estimation software that relies on creating
an underlying 3D model of a person from an image. SMPLify-X \cite{pavlakos2019expressive} an extended version of the SMPL model \cite{loper2015smpl} was employed.
SMPL is a principal components model of laser scans of human bodies \cite{Loper2015}. 
From 2D detected keypoints,  3D body parameters (pose $\theta$, shape $\beta$, and expression $\psi$) were estimated 
 by minimizing the difference between the 2D keypoints
and the 3D keypoints re-projected into the image plane.  Weight estimates for above- and below-average sized bodies were poor. Including 2D silhouette information improved estimates.

In another approach, a data-driven strategy was used to test 
scale agnostic body shape classification for 
large numbers of 3D body models synthesized from sampled SMPL $\beta$ 
parameters \cite{Thakkar_2022_CVPR}. Equivalence classes  (19 for female bodies, and 15 for male bodies---manually narrowed to 12 categories each) were found.
The model was evaluated for reliability in
estimating  height, weight, and non-metric categorization of 3D shapes using a scale-agnostic
measure of body shape. Although the approach improves on previous efforts \cite{thakkar2021feasibility}, accurate body-shape identification using 3D models created from a single, reference-free image remains challenging.

\textbf{Linguistic Descriptions for Body Synthesis.}
Our use of linguistic
 descriptors as a body biometric was inspired by  
two  studies \cite{hill2016creating, streuber2016body}. Both   demonstrate that it is possible to reconstruct a perceptually and metrically  accurate three-dimensional model of a person's body by learning the mapping between 30 linguistic attributes (see Table \ref{descriptors}) and the $\beta$ coefficients from the SMPL model.
We reasoned that if these 30 linguistic attributes
were useful in reconstructing an accurate body model, they might also serve as a useful representation of body shape that could be used to identify, or at least classify bodies by shape.

The problem we undertake requires learning a
mapping from 2D images of a body to the linguistic descriptor terms. 
The goal of the present work was to use the descriptor-based representation as part of a core ResNet101 model that could be refined by explicitly using transfer learning to train the model for identification.

\section{Experiments}



We conducted experiments to test the
role of linguistic features in identification of people by body shape.  A model trained to predict a word-based description (see Table \ref{descriptors})
of a human body served as the core ResNet model for the linguistic model. The model was then trained with transfer learning to identify 
people by body.  This linguistic  model was compared to an otherwise identical model without linguistic training.  

\begin{table*}
\begin{center}
\begin{tabular}{|l|c|c|c|c|c|c|c|}
\hline
Training & IDs & Images &\multicolumn{5}{ c |}{Distance \& Pitch (UAV)} \\
\hline\hline
BRS1 & 158 & 46834 &100m & 200m & 400m & 500m & UAV \\
BRS1.1 & 54 & 13288 & 100m & 300m & 400m & 500m & UAV \\
BRS2 & 195 & 51134 &100m & 270m & 370m & 800m & 1000m\\
BRS3 & 170 & 88573 & 100m& 200m & 300m & 500m & \\
\hline
\end{tabular}
\end{center}
\caption{\small Training data subsets indicating number of identities and distances in meters. UAV refers to image frames in video taken from an unmanned aerial vehicle.}
\label{BRS_info}
\vskip -0.4cm
\end{table*}
\subsection{Datasets}\label{datasets}

\textbf{Linguistic Training}.
Training of the linguistic core was done with annotations of 
identities using single frames extracted
from two data sets:
the Human Identification (HumanID) \cite{OToole_2005},
and the Multiview Extended Video with Activities (MEVA) datasets \cite{Corona_2021_WACV}.
For the former,
three types of videos were sampled: a.) people approaching from a distance, b.) people passing perpendicular to a camera, and c.) people viewed from an approximately 45$\degree$ raised pitch. Annotations were collected on each of the 297 identities available in the dataset from
20 annotators.  Specifically, annotations were collected from (all/most) frames
in which a body could be detected, 
yielding a total of approximately (250K) frames.
These frames were
annotated with the 30 linguistic descriptors listed in Table \ref{descriptors}. 
Descriptors for each identity were averaged across  annotators and the averages were used for training. The network learned to produce the descriptors from the images.

The MEVA dataset \cite{Corona_2021_WACV} is a  very-large-scale dataset for human activity recognition. It contains over 9,300 hours of untrimmed video, loosely scripted to include a diverse range of activities and background scenes. There are 37 activity types performed by approximately 100 actors in scripted scenarios.  
Human ratings were gathered on all 158 available identities in the dataset.  Because no explicit identity labels were provided in 
the MEVA
dataset, images were cropped out of the 7,000 available videos and 
annotated by between 11--15 annotators.  Again, the network 
was trained to produce the descriptors from the images.

\textbf{Identification Transfer Training.} To train and evaluate models, we used the BRIAR dataset \cite{cornett2023expanding}, which is divided into a training (BRIAR Research Set, BRS) and a test set (BRIAR Test Set, BTS). These sets were issued in multiple releases, which included increasing numbers of identities.
The BRS was used for identification transfer training, and a 
curated subset of BTS was used for testing (see 
Section \ref{sec:evaluation} for details of this dataset).  

\begin{figure}[htbp]
\begin{center}
   \includegraphics[width=3in]{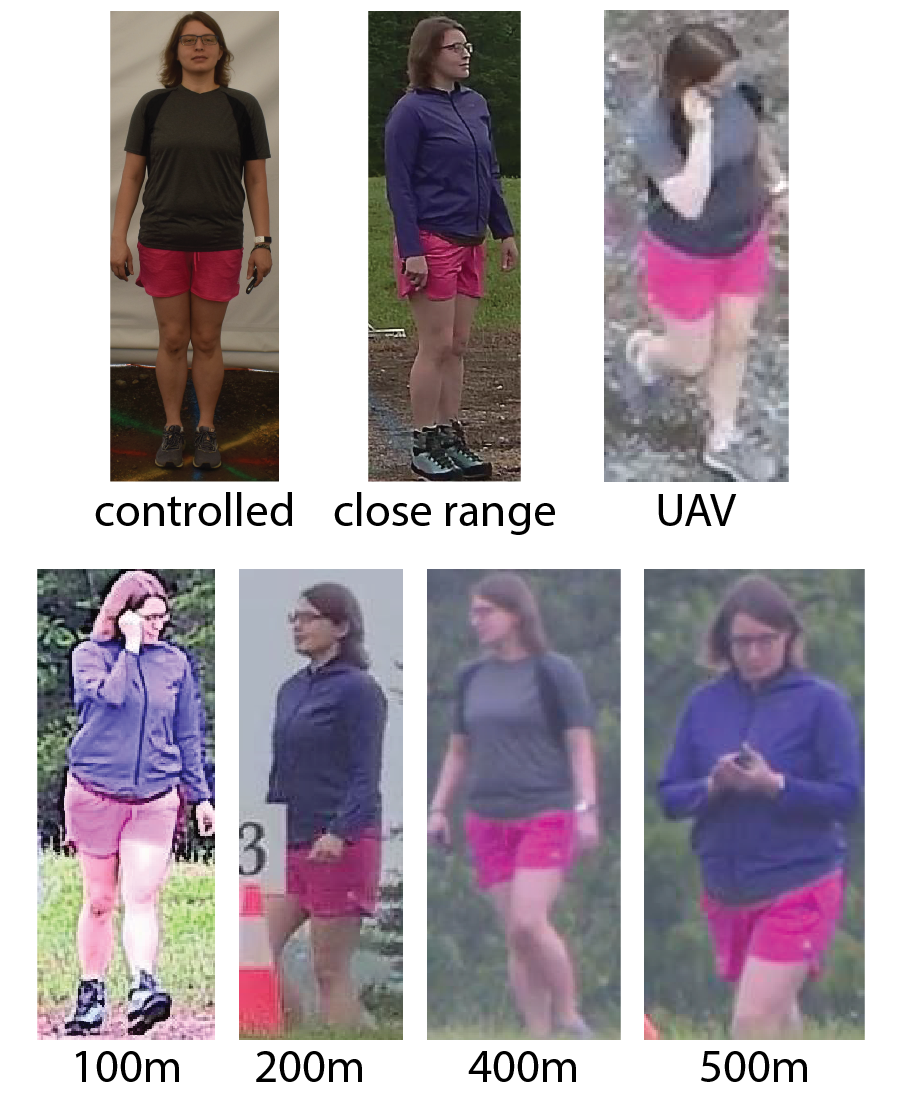}
\end{center}
   \caption{Example images of one identity at each distance in the BRS1 dataset.}
\label{fig:distances}
\vskip -0.4cm
\end{figure}

For training, we utilized BRS1--3 (See Table \ref{BRS_info}).
For each subject, there are indoor (controlled) and outdoor (field) sequences. In the controlled set, cameras captured walking videos of two types: structured walking and random walking. In addition to the walking types, there were also standing videos. All distances except for the close-range set had one viewing angle, whereas the close-range set had three cameras capturing the same position from different yaw angles: 0$\degree$, 30$\degree$, and 50$\degree$. All sequences involved two clothing settings.
Example identities appear in Figure \ref{fig:BRS_examples}. Examples of images
taken at different distance conditions and from UAV appear in 
Figure \ref{fig:distances}.

\begin{figure}[t]
\begin{center}
   \includegraphics[width=\linewidth]{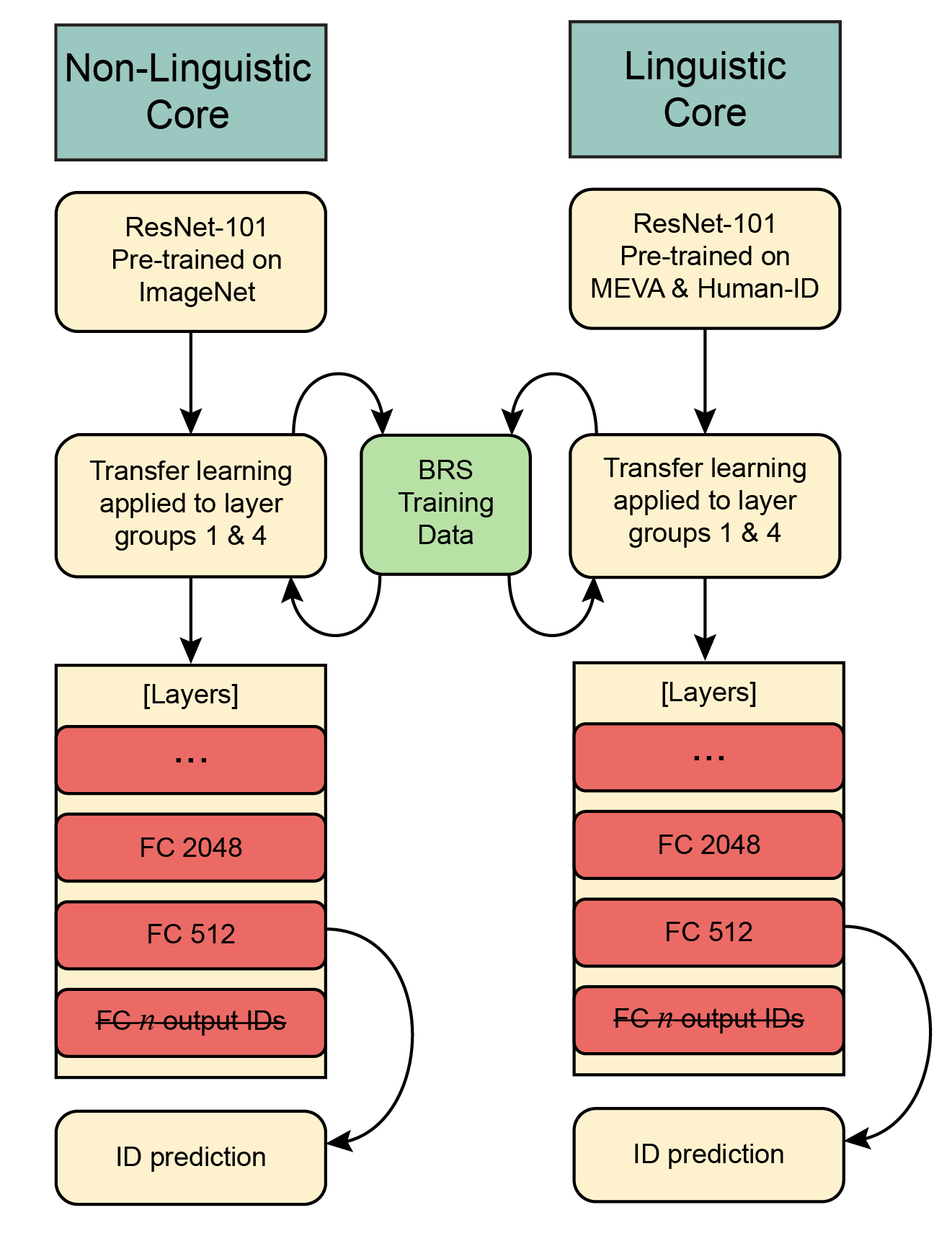}
\end{center}
   \caption{Model architecture with image core (left) and linguistic core (right). Both are trained for identification using image frames from BRS. ``Layer groups'' refer to network subdivisions used by the code implementation (see text).}
\label{fig:model_schematic}
\vskip -0.4cm
\end{figure}

\subsection{Model Components}
\vskip -0.2cm
We created two identity-trained models that 
differed
only in the base model to which identity transfer learning was applied. We refer to these as the Linguistic Core ResNet Identity Model (LCRIM) and the
Non-Linguistic Core ResNet Identity Model (NLCRIM). A fusion model was created by combining the results of the LCRIM and NLCRIM models. Specifically, cosine similarities for all possible probe-gallery pairs were computed for both the LCRIM and NLCRIM models. The average of the two cosines served as the representation for the fusion model.

\textbf{Linguistic and Non-linguistic Core Training}.  
The {\it linguistic core} model derives from a ResNet-101 architecture. 
Specifically, the model consists of a base ResNet 101 \cite{he2015deep} pre-trained with ImageNet \cite{russakovsky2015imagenet}. Transfer learning
was applied to the base model to map from image frames in the MEVA and
HumanID datasets (see Section \ref{datasets}) to the 30 linguistic body
attributes (see Table \ref{descriptors}). 
The architecture of the final model consisted of the ResNet
base with an appended encoder (2048 $\rightarrow$ 512 $\rightarrow$ 64 
$\rightarrow$ 16) 
to a decoder 
(16 $\rightarrow$ 24 $\rightarrow$ 30). 
The {\it non-linguistic core} model was simply the  base ResNet 101 \cite{he2015deep} pre-trained with ImageNet \cite{russakovsky2015imagenet}.



\begin{figure*}[t]
\begin{center}
   \includegraphics[width=0.45\linewidth]{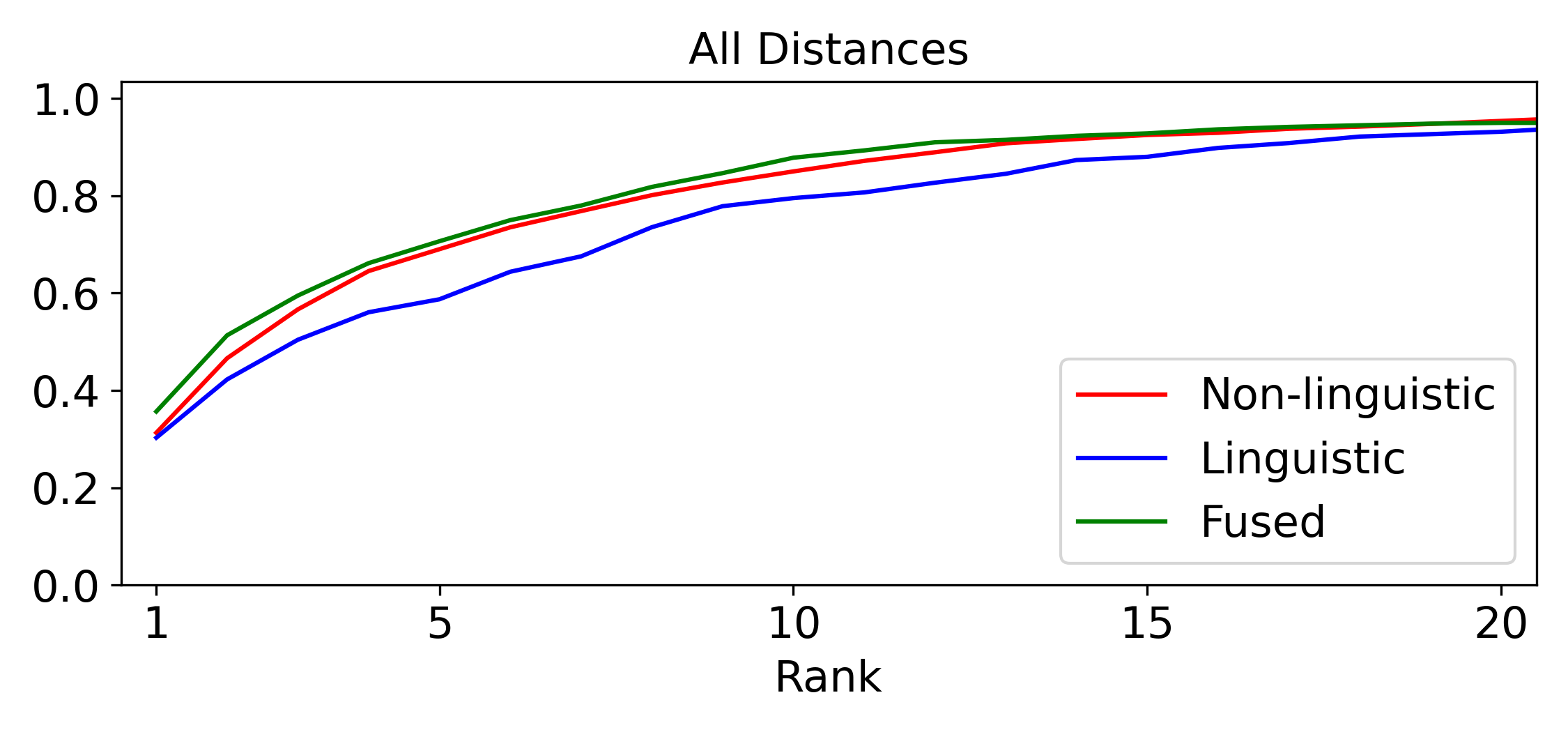}\\
   \includegraphics[width=0.45\linewidth]{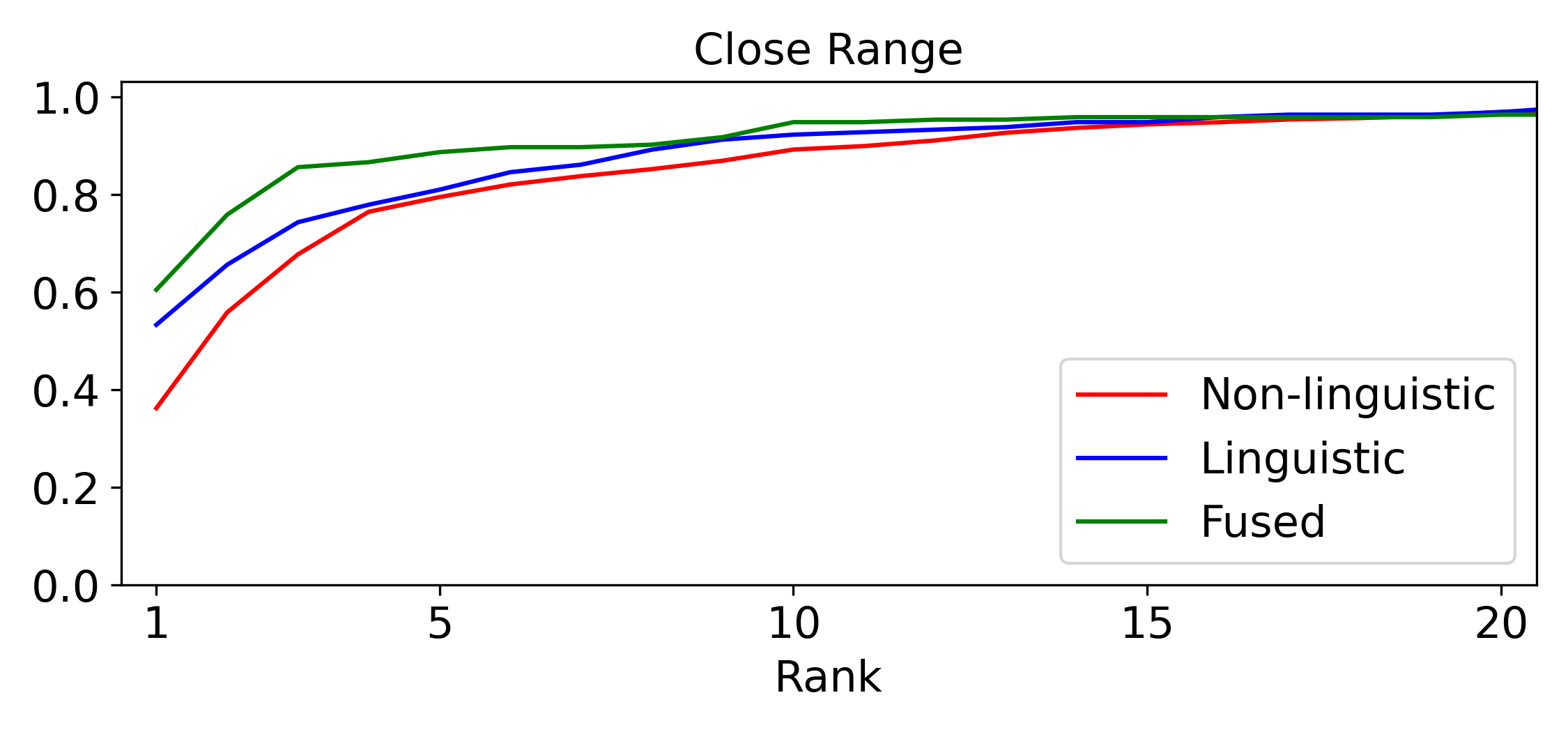}
   \includegraphics[width=0.45\linewidth]{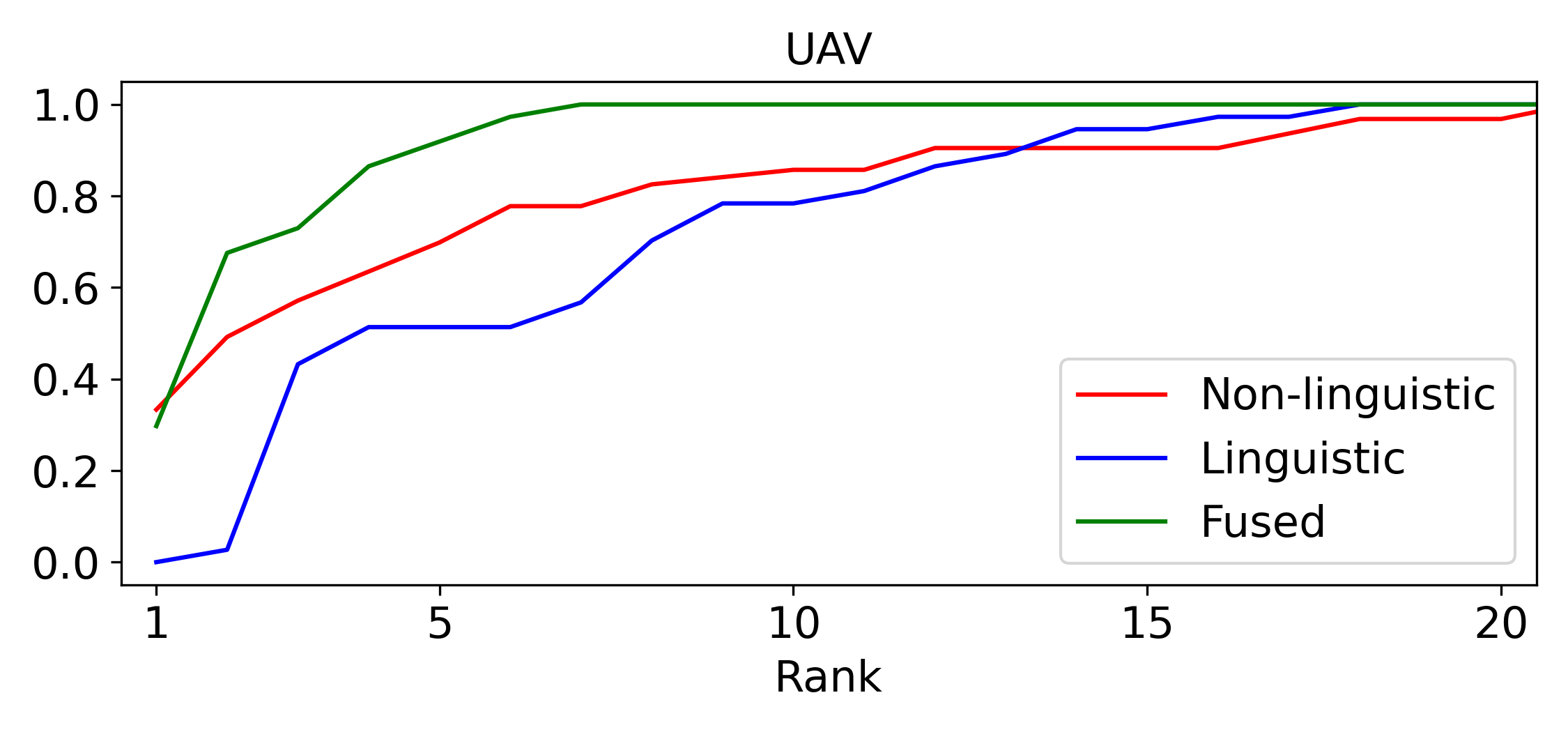}\\   \includegraphics[width=0.45\linewidth]{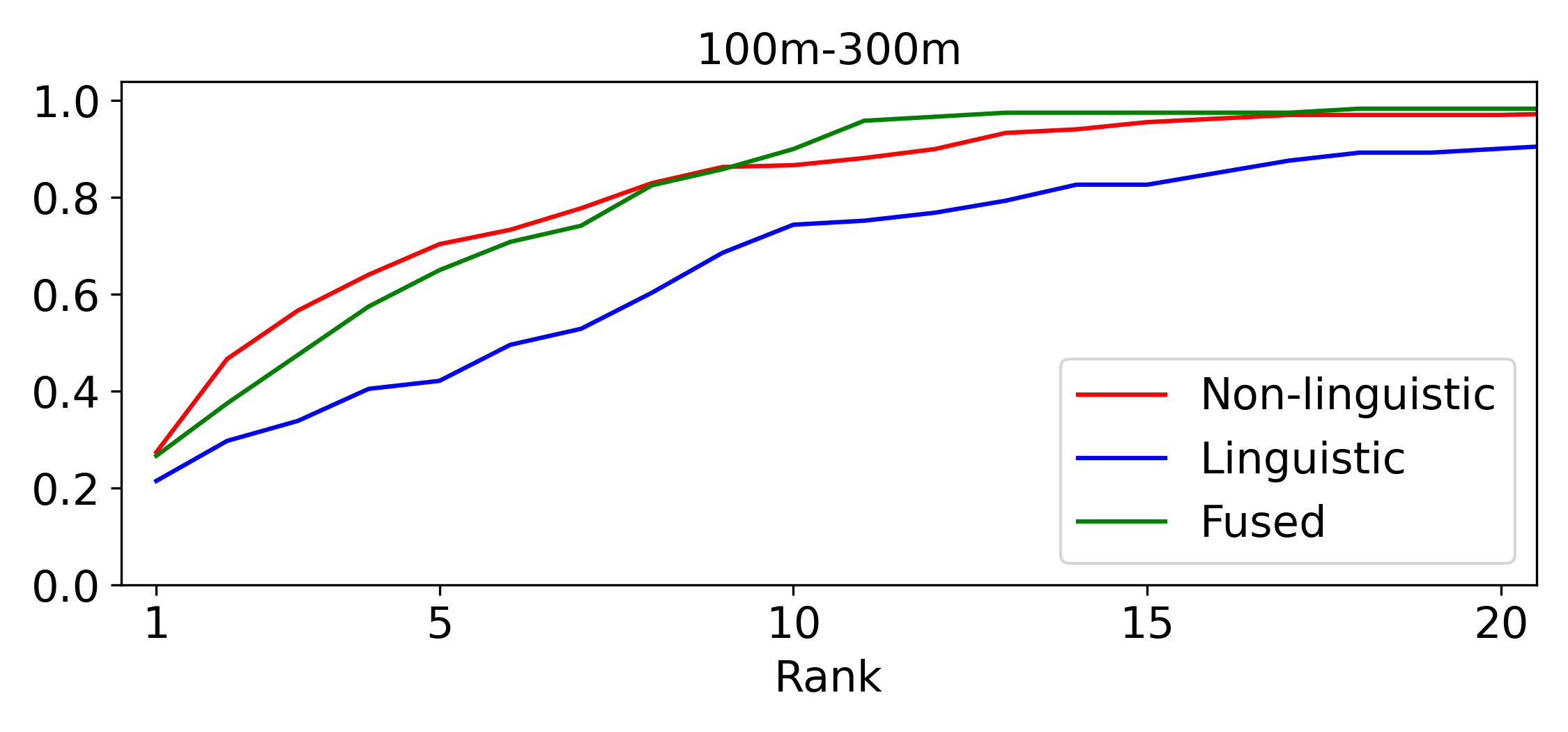}
   \includegraphics[width=0.45\linewidth]{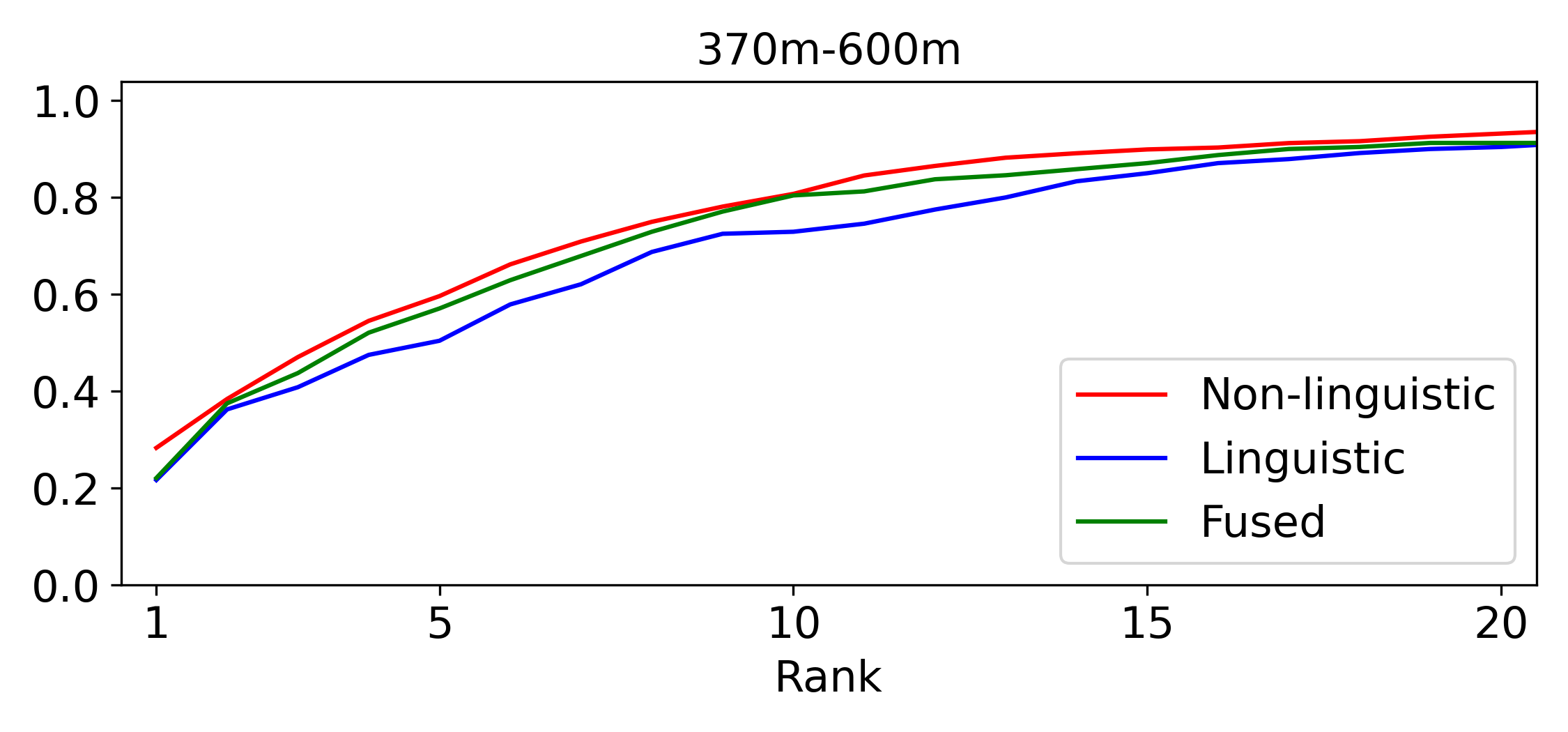}
\end{center}
   \caption{Cumulative Match Characteristic (CMC) curves comparing the performance of the Linguistic (LCRIM) and Non-Linguistic Core ResNet Identity models (NLCRIM), plotted with the fusion of the models. At close range, the linguistic model performed more accurately across all ranks. The non-linguistic model fared better at intermediary distances.  Notably, fusion of the two model embeddings was helpful in most cases, regardless of whether the linguistic or non-linguistic model performed more accurately.}
\label{fig:CMC}
\vskip -0.4cm
\end{figure*}

\begin{table}
\begin{center}
\begin{tabular}{|l|c|c|c|c|c|c|}
\hline
Distance & Gallery IDs & Probe IDs & Probe Images \\
\hline\hline
Close Range & 60 & 95 & 1353 \\
100m--300m & 60 & 86 & 638 \\
370m--600m & 60 & 92 & 1354 \\
UAV & 60 & 17 & 204 \\

\hline
\end{tabular}
\end{center}
\caption{BTS dataset used for model evaluation as a function of distance/view. In this dataset, there are more unique probe identities
($n=100$) than unique gallery identities ($n=60$) making this an open-set test.}
\label{Table:BTS}
\vskip -0.4cm 
\end{table}

\textbf{Identification Transfer Learning for Bodies.}
Both networks consist of core models (linguistic, non-linguistic) that were altered via transfer learning for identification (i.e., separating individual people by identity). This learning was applied to the first and fourth network layer groups. Here, ``layer groups'' refer to the five subsections of the ResNet architecture used in the PyTorch implementation and detailed in Table 2 of \cite{he2015deep}. These layer groups each consist of multiple residual blocks.
To adapt the ResNet 101 for body identification,
identity training was implemented using still images for each identity, as well as image frames extracted from the BRS1--3 videos. Image frame extraction was done by partitioning video frames into equal-length groupings and randomly selecting 5 frames across the video partitions. Image crops were then generated with a pre-trained Inception-Net-V2 object detector \cite{szegedy2017inception}. 
Low-confidence crops were discarded to assure high-quality crops. This frame selection method generated a total of 199,829 images of 577 
identities
across the distances and views available in BRS (see Table \ref{BRS_info}). 

We pre-processed the dataset by resizing the images to 128x256 pixels (retaining aspect ratio) and normalizing the pixel values. The  dataset was split into training and validation sets in a ratio of 80:20. The model was trained using the cross-entropy loss function and the Adam optimizer with a starting learning rate of 0.00005 and momentum of 0.9. We used flipping and rotation as data augmentation techniques. The end part of the model architecture was an Adaptive Average Pooling layer, 
followed by a fully connected layer with 2048 input features and 512 output features. A Parametric Rectified Linear Unit (PReLU) activation function was then applied. Lastly, a sequential block containing a fully connected layer mapped the 512 input features to the 577 identities on which the model was trained.

\begin{figure*}[t]
\begin{center}
   \includegraphics[width=0.4\linewidth]{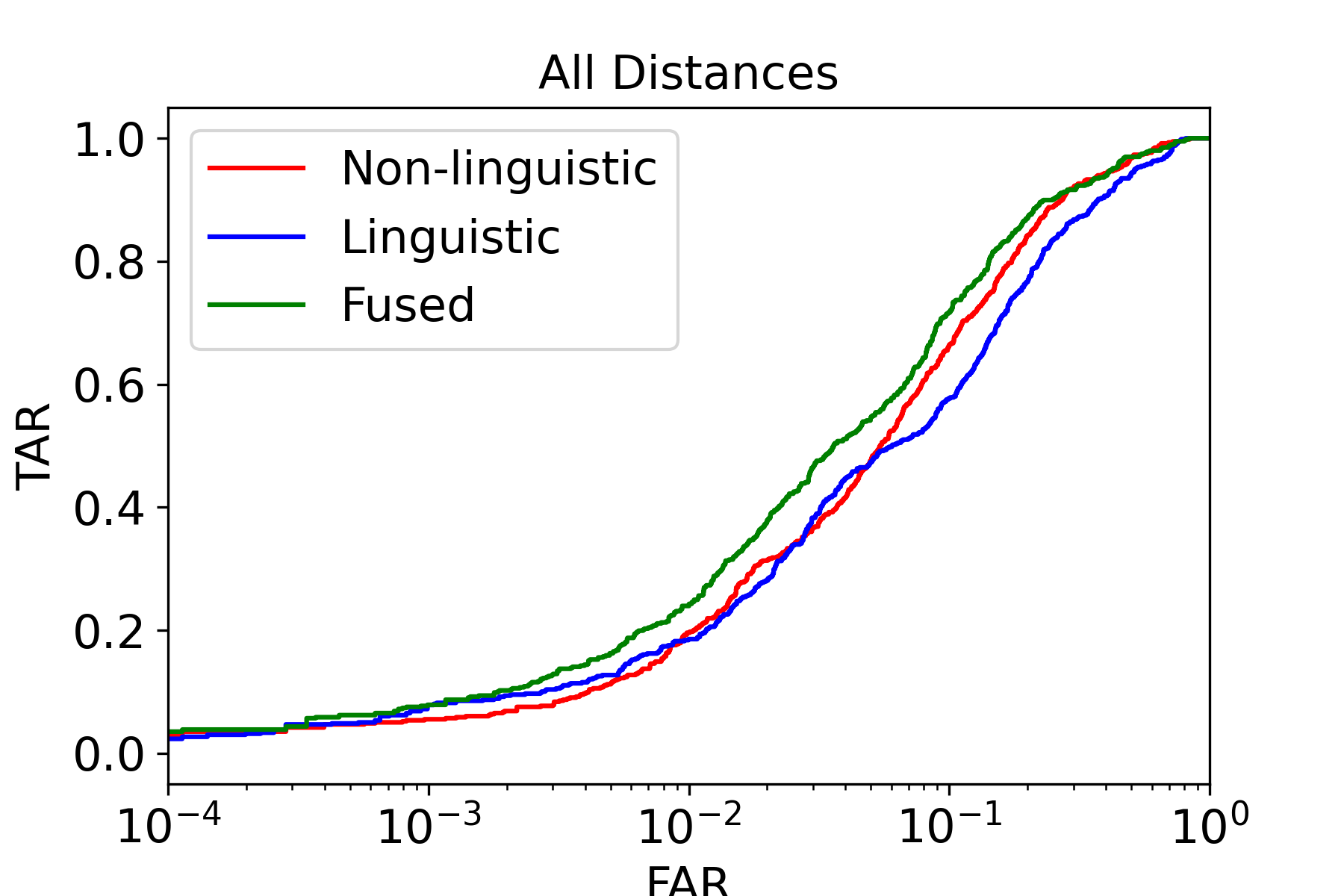}\\
   \includegraphics[width=0.4\linewidth]{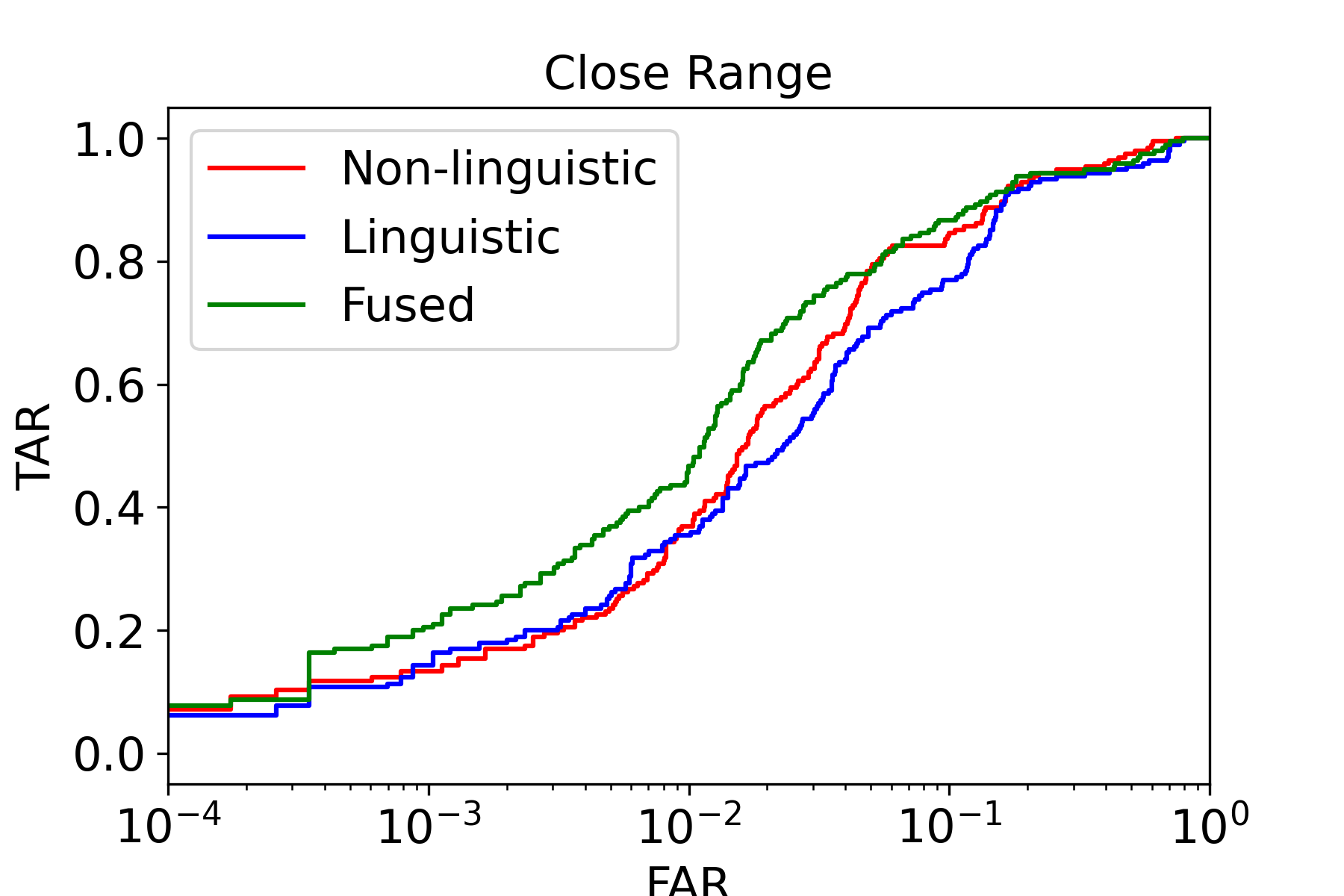}
   \includegraphics[width=0.4\linewidth]{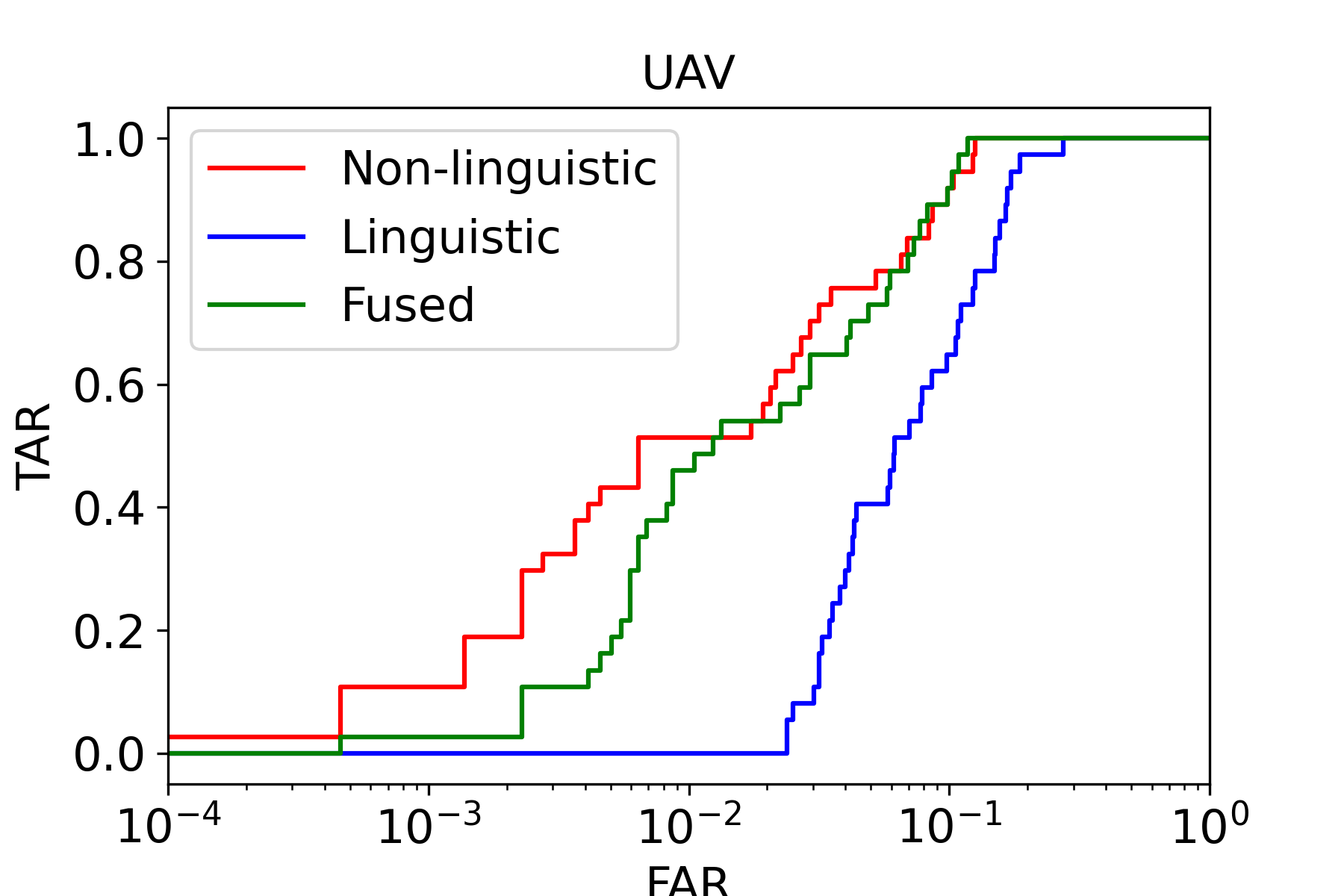}\\   \includegraphics[width=0.4\linewidth]{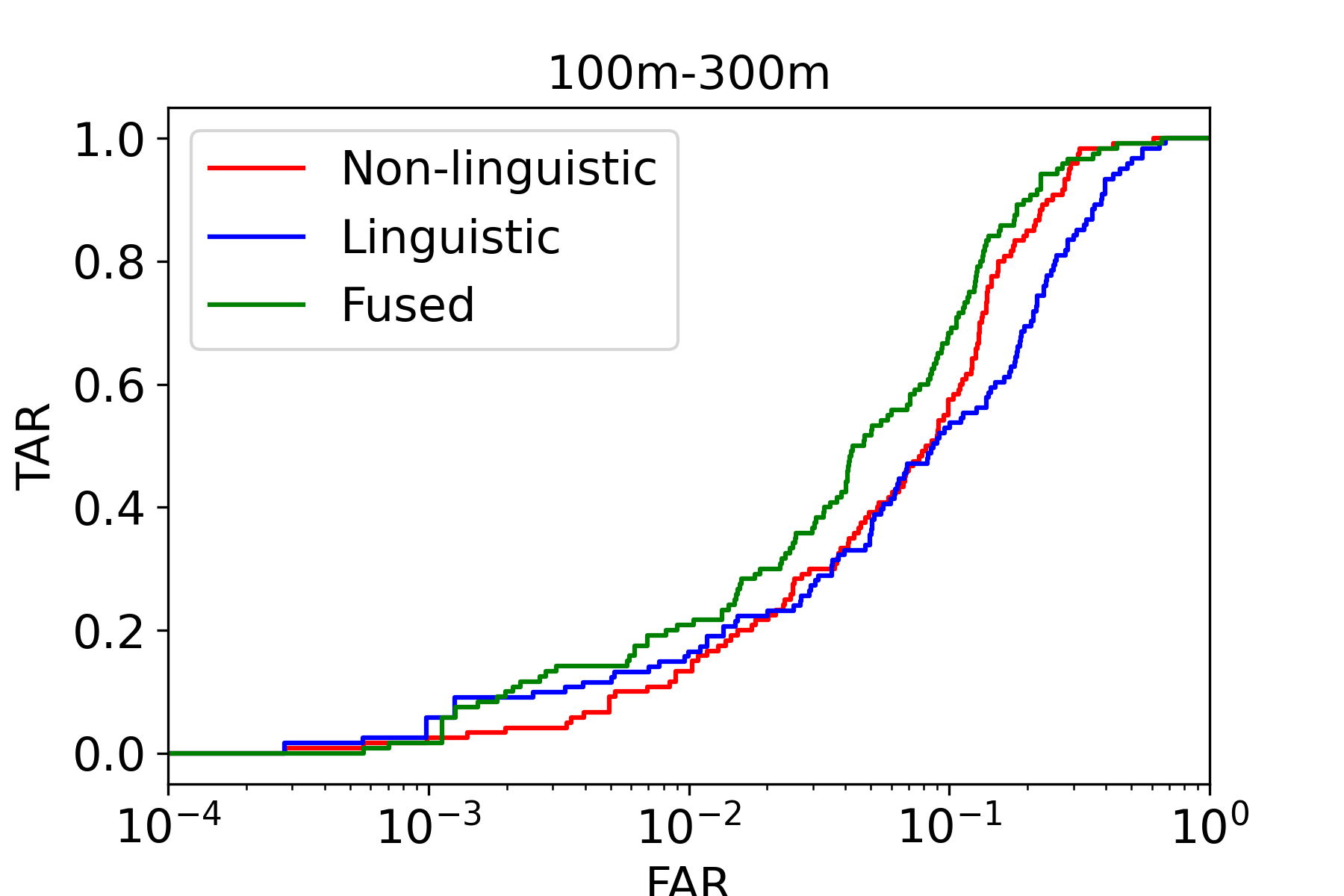}
      \includegraphics[width=0.4\linewidth]{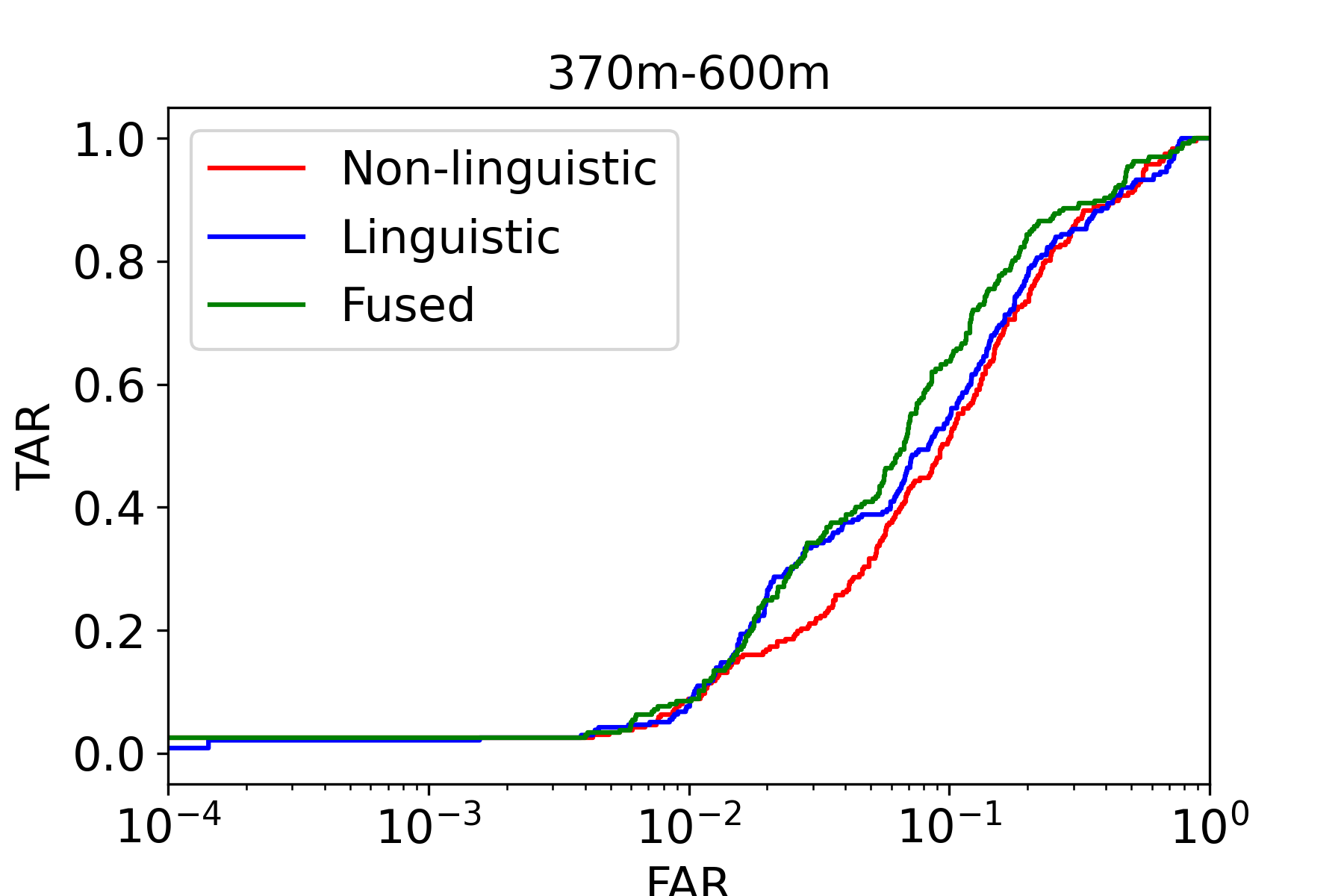}
\end{center}
   \caption{Receiver Operating Characteristic (ROC) curves comparing the performance of the Linguistic (LCRIM) and Non-Linguistic Core ResNet Identity models (LCRIM), plotted with the fusion of the models. Although the non-linguistic model fared better in all cases, the fusion of the two model embeddings improved performance for all cases, with the exception of the UAV. (True accept rate, TAR; False accept rate, FAR)}
\label{fig:verification}
\vskip -0.4cm
\end{figure*}

\subsection{Model Evaluation}\label{sec:evaluation}

\textbf{Test Data}.  
Test data from the BTS are summarized in Table \ref{Table:BTS}. The gallery consisted of both images and videos. The probe set consisted of videos only. In all cases, matched-identity gallery and probe items 
wore different clothes.
For gallery items, images were processed through both networks to produce 
a feature-embedding for each network. These embeddings  were extracted from the penultimate layer  of each network (512 units).
For each gallery identity, the feature vectors 
from the images and videos were averaged to produce a single representation for each unique identity.



\begin{table*}
\begin{center}
\resizebox{\textwidth}{!}{%
\begin{tabular}{|l|c c c|c c c|c c c|c c c|c c c|}
\hline
Model & \multicolumn{3}{c|}{All Distances} & \multicolumn{3}{c|}{Close Range} & \multicolumn{3}{c|}{UAV} & \multicolumn{3}{c|}{100m--300m} & \multicolumn{3}{c|}{370m--600m} \\
\hline\hline
NLCRIM & 0.31 & 0.85 & \textbf{0.95} & 0.36 & 0.89 & \textbf{0.97} & \textbf{0.33} & 0.86 & 0.97 & \textbf{0.27} & 0.87 & 0.97 & \textbf{0.28} & \textbf{0.81} & \textbf{0.93} \\
LCRIM & 0.30 & 0.80 & 0.93 & 0.53 & 0.92 & \textbf{0.97} & 0.00 & 0.78 & \textbf{1.00} & 0.21 & 0.74 & 0.90 & 0.22 & 0.73 & 0.90 \\
Fused & \textbf{0.36} & \textbf{0.88} & \textbf{0.95} & \textbf{0.61} & \textbf{0.95} & 0.96 & 0.30 & \textbf{1.00} & \textbf{1.00} & \textbf{0.27} & \textbf{0.90} & \textbf{0.98} & 0.22 & 0.80 & 0.91 \\
\hline
\end{tabular}
}
\end{center}
\caption{Performance as the proportion of probe items correctly matched to their gallery identity at rank 1, rank 10, and rank 20. Fusion improves most categories of items.}
\label{Table:ranks}
\vskip -0.4cm
\end{table*}

For both the gallery and probe videos, every sixth frame was extracted. These frames were processed through both networks to produce feature embeddings, again extracted from
the penultimate layer of 512 units.
A single representation of the video was made by averaging
the features for the extracted frames.

\textbf{Quantitative Evaluation Procedure}. Model testing was implemented as 
a 1-to-1 verification  between gallery and probe items, with cosine as the similarity metric. Performance was analyzed in two ways. Cumulative 
Match Characteristic (CMC) curves show the proportion of correctly matched items
as a function of the rank of the match. 
Receiver Operating Characteristic (ROC) curves show the 
true accept rate as a function of the false accept rate.
This latter curve is particularly useful for examining model performance in the open set case, where the number of unique gallery identities is less than  unique probes identities.

\textbf{Ranked Match Results}. The CMC results appear in Figure \ref{fig:CMC} and show data for the three models, NLCRIM, LCRIM, and Fused. Each plot shows CMC curves displayed by the condition: Close Range, UAV, 100m--300m, and 370m--600m, and averaged across all conditions (All Distances). Table \ref{Table:ranks}
gives the proportion of correct identifications at Rank 1, 10, and 20 by distance condition and as summarized over all distances.  

Identification performance was surprisingly accurate. As the table shows, 
across all distances in the fused model, 36\% of the 
probe items with a matched identity in the
gallery were matched at Rank 1, 88\% were matched by Rank 10, and 95\% were matched by Rank 20. 
Figure \ref{fig:CMC} shows that the Fused model was slightly, but consistently, superior to the NLCRIM and LCRIM models at all ranks. This indicates 
that models with and without a linguistic core contain 
complementary information for body identification that can be combined to improve performance. 
That conclusion is supported by the rank data in Table \ref{Table:ranks}.  
At Rank 1, 10, and 20, fusion yields the best performance obtained at these ranks.  

Dividing the data by category, 
Figure \ref{fig:CMC} and Table \ref{Table:ranks} show that the LCRIM model performed more accurately 
than NLCRIM in the close-range condition, whereas
NLCRIM was more accurate in the other conditions
(100--300m, 370--600m, UAV). The fused model fared better than both the LCRIM and NLCRIM
for the close range and UAV data.
For distances between 100 and 300m, fusion dominated only at ranks greater than 10.
Not surprisingly, the poorest performance for all models was found for
distances between 370 and 600m. In this condition, the NLCRIM, rather than the Fused model, performed best. 

The UAV condition is of special interest. Despite the strong superiority of the NLCRIM over the LCRIM, and the relatively weak performance of the LCRIM, fusion improved performance by a large margin at all ranks---with 
the fused model reaching 100\% correct matches by Rank 10. This indicates the utility of combining these two sources of information about bodies with extreme overhead pitch.  


\textbf{Verification Rate Results}. The ROC results appear in Figure \ref{fig:verification} and provide an evaluation of the models' performance across various operating points. The ROC is especially important because it includes false match error 
for probe items that do not have a matching gallery entry. 
These results complement the CMC analysis but offer additional information about the challenges of the open set problem.

To begin, as for the CMC data, the ROC results show a small, but consistent, benefit from fusing the LCRIM and NLCRIM models across a wide swath of false accept rates. 
The  advantage of fusion was consistent across false accept rates for the close range and 100--300m conditions, and across some but not all false accept rates in the 370--600m condition. Contrary to the fusion advantage seen with the CMC measure, using the ROC measure, fusion did not benefit
the UAV category. Instead, the NLCRIM model performed best.  

In examining the NLCRIM and LCRIM models by themselves, neither showed a clear and consistent advantage over the other at all ranks. This further emphasizes the potential for a benefit of fusion across  different conditions and different operating points of the model.

\section{Discussion}

A key challenge in person identification using body shape is the considerable variability in appearance due to changes in clothing, viewpoint, and distance. The limited availability of high-quality datasets for training body recognition algorithms exacerbates these challenges. We addressed these issues by employing a combination of object recognition strategies and linguistic body descriptions, with the idea
that linguistic descriptions might be more robust to changes in appearance and might instead rely more fundamentally on body shape for identification.

The approach began with a deep network  trained to predict linguistic descriptors of body shape. Transfer learning was applied to train this core network to identify people based on body images. We compared the performance of this linguistic model  with that of an identical model without linguistic training. The models were evaluated with a diverse test set comprised of images and videos taken over a wide variety of distances and from an elevated pitch. 

The linguistic  and non-linguistic models performed surprisingly well  
at body identification.
This suggests that there may be
more identity-diagnostic information in bodies than  previously assumed. 
As expected, body identification was best at close distances.
but remained relatively robust  with hundreds of meters of distance and
from elevated pitch (UAV).
Given that body shape is not a unique biometric, this result has important implications. 
Person identification can be critical in cases where the face is not visible or resolvable, the possibility of relying on body information, either in isolation or in combination with gait. 

The utility of fusing different kinds of 
information from static bodies for body identification  was evident.
A fusion of the linguistic and non-linguistic
models improved body identification in most cases. The proposed method operated effectively in demanding testing conditions, including at a distance, with overhead viewing, and with mismatched clothing. This highlights the potential of the approach for applications in security and surveillance, where person identification often relies on distant and obstructed views of the subject.

In conclusion, we demonstrate the feasibility of  body shape for person identification, using a novel approach that leverages linguistic body descriptions with an object classification approach. The fusion of linguistic and non-linguistic information can improve body identification, offering a promising direction for future research and practical applications in challenging viewing conditions.








{\small
\bibliographystyle{ieee}
\bibliography{Body_IJCB/references}

\begin{thebibliography}{10}\itemsep=-1pt

\bibitem{cornett2023expanding}
D.~Cornett, J.~Brogan, N.~Barber, D.~Aykac, S.~Baird, N.~Burchfield, C.~Dukes,
  A.~Duncan, R.~Ferrell, J.~Goddard, et~al.
\newblock Expanding accurate person recognition to new altitudes and ranges:
  The briar dataset.
\newblock In {\em Proceedings of the IEEE/CVF Winter Conference on Applications
  of Computer Vision}, pages 593--602, 2023.

\bibitem{Corona_2021_WACV}
K.~Corona, K.~Osterdahl, R.~Collins, and A.~Hoogs.
\newblock Meva: A large-scale multiview, multimodal video dataset for activity
  detection.
\newblock In {\em Proceedings of the IEEE/CVF Winter Conference on Applications
  of Computer Vision (WACV)}, pages 1060--1068, January 2021.

\bibitem{geirhos2018imagenet}
R.~Geirhos, P.~Rubisch, C.~Michaelis, M.~Bethge, F.~A. Wichmann, and
  W.~Brendel.
\newblock Imagenet-trained cnns are biased towards texture; increasing shape
  bias improves accuracy and robustness.
\newblock {\em arXiv preprint arXiv:1811.12231}, 2018.

\bibitem{godil2003human}
A.~Godil, P.~Grother, and S.~Ressler.
\newblock Human identification from body shape.
\newblock In {\em Fourth International Conference on 3-D Digital Imaging and
  Modeling, 2003. 3DIM 2003. Proceedings.}, pages 386--392. IEEE, 2003.

\bibitem{hahn2016dissecting}
C.~A. Hahn, A.~J. O'Toole, and P.~J. Phillips.
\newblock Dissecting the time course of person recognition in natural viewing
  environments.
\newblock {\em British Journal of Psychology}, 107(1):117--134, 2016.

\bibitem{he2015deep}
K.~He, X.~Zhang, S.~Ren, and J.~Sun.
\newblock Deep residual learning for image recognition, 2015.

\bibitem{hill2016creating}
M.~Q. Hill, S.~Streuber, C.~A. Hahn, M.~J. Black, and A.~J. O’Toole.
\newblock Creating body shapes from verbal descriptions by linking similarity
  spaces.
\newblock {\em Psychological science}, 27(11):1486--1497, 2016.

\bibitem{kale2004identification}
A.~Kale, A.~Sundaresan, A.~Rajagopalan, N.~P. Cuntoor, A.~K. Roy-Chowdhury,
  V.~Kruger, and R.~Chellappa.
\newblock Identification of humans using gait.
\newblock {\em IEEE Transactions on image processing}, 13(9):1163--1173, 2004.

\bibitem{loper2015smpl}
M.~Loper, N.~Mahmood, J.~Romero, G.~Pons-Moll, and M.~J. Black.
\newblock Smpl: A skinned multi-person linear model.
\newblock {\em ACM transactions on graphics (TOG)}, 34(6):1--16, 2015.

\bibitem{Loper2015}
M.~Loper, N.~Mahmood, J.~Romero, G.~Pons-Moll, and M.~J. Black.
\newblock Smpl: A skinned multi-person linear model.
\newblock {\em ACM Trans. Graph.}, 34(6), nov 2015.

\bibitem{myronenko2010point}
A.~Myronenko and X.~Song.
\newblock Point set registration: Coherent point drift.
\newblock {\em IEEE transactions on pattern analysis and machine intelligence},
  32(12):2262--2275, 2010.

\bibitem{OToole_2005}
A.~O'Toole, J.~Harms, S.~Snow, D.~Hurst, M.~Pappas, J.~Ayyad, and H.~Abdi.
\newblock A video database of moving faces and people.
\newblock {\em IEEE Transactions on Pattern Analysis and Machine Intelligence},
  27(5):812--816, 2005.

\bibitem{pavlakos2019expressive}
G.~Pavlakos, V.~Choutas, N.~Ghorbani, T.~Bolkart, A.~A. Osman, D.~Tzionas, and
  M.~J. Black.
\newblock Expressive body capture: 3d hands, face, and body from a single
  image.
\newblock In {\em Proceedings of the IEEE/CVF conference on computer vision and
  pattern recognition}, pages 10975--10985, 2019.

\bibitem{rice2013unaware}
A.~Rice, P.~J. Phillips, V.~Natu, X.~An, and A.~J. O’Toole.
\newblock Unaware person recognition from the body when face identification
  fails.
\newblock {\em Psychological Science}, 24(11):2235--2243, 2013.

\bibitem{rice2013role}
A.~Rice, P.~J. Phillips, and A.~O'Toole.
\newblock The role of the face and body in unfamiliar person identification.
\newblock {\em Applied Cognitive Psychology}, 27(6):761--768, 2013.

\bibitem{robbins2012effects}
R.~A. Robbins and M.~Coltheart.
\newblock The effects of inversion and familiarity on face versus body cues to
  person recognition.
\newblock {\em Journal of Experimental Psychology: Human Perception and
  Performance}, 38(5):1098, 2012.

\bibitem{russakovsky2015imagenet}
O.~Russakovsky, J.~Deng, H.~Su, J.~Krause, S.~Satheesh, S.~Ma, Z.~Huang,
  A.~Karpathy, A.~Khosla, M.~Bernstein, et~al.
\newblock Imagenet large scale visual recognition challenge.
\newblock {\em International journal of computer vision}, 115:211--252, 2015.

\bibitem{streuber2016body}
S.~Streuber, M.~A. Quiros-Ramirez, M.~Q. Hill, C.~A. Hahn, S.~Zuffi,
  A.~O'Toole, and M.~J. Black.
\newblock Body talk: Crowdshaping realistic 3d avatars with words.
\newblock {\em ACM Transactions on Graphics (TOG)}, 35(4):1--14, 2016.

\bibitem{szegedy2017inception}
C.~Szegedy, S.~Ioffe, V.~Vanhoucke, and A.~Alemi.
\newblock Inception-v4, inception-resnet and the impact of residual connections
  on learning.
\newblock In {\em Proceedings of the AAAI conference on artificial
  intelligence}, volume~31, 2017.

\bibitem{thakkar2021feasibility}
N.~Thakkar and H.~Farid.
\newblock On the feasibility of 3d model-based forensic height and weight
  estimation.
\newblock In {\em Proceedings of the IEEE/CVF Conference on Computer Vision and
  Pattern Recognition}, pages 953--961, 2021.

\bibitem{Thakkar_2022_CVPR}
N.~Thakkar, G.~Pavlakos, and H.~Farid.
\newblock The reliability of forensic body-shape identification.
\newblock In {\em Proceedings of the IEEE/CVF Conference on Computer Vision and
  Pattern Recognition (CVPR) Workshops}, pages 44--52, June 2022.

\bibitem{ye2021deep}
M.~Ye, J.~Shen, G.~Lin, T.~Xiang, L.~Shao, and S.~C. Hoi.
\newblock Deep learning for person re-identification: A survey and outlook.
\newblock {\em IEEE transactions on pattern analysis and machine intelligence},
  44(6):2872--2893, 2021.

\bibitem{yovel2016recognizing}
G.~Yovel and A.~J. O’Toole.
\newblock Recognizing people in motion.
\newblock {\em Trends in cognitive sciences}, 20(5):383--395, 2016.

\end{thebibliography}
}

\section{Acknowledgements}
This research is based upon work supported in part by
the Office of the Director of National Intelligence (ODNI),
Intelligence Advanced Research Projects Activity (IARPA),
via [2022-21102100005]. The views and conclusions contained herein are those of the authors and should not be
interpreted as necessarily representing the official policies,
either expressed or implied, of ODNI, IARPA, or the U.S.
Government. The US. Government is authorized to reproduce
and distribute reprints for governmental purposes notwithstanding any copyright annotation therein.

\end{document}